\documentclass[twocolumn]{svjour3}          
\smartqed  

\usepackage{graphicx}
\usepackage{times}
\usepackage{epsfig}
\usepackage{subfigure}
\usepackage{amsmath}
\usepackage{amssymb}
\usepackage{algorithm}
\usepackage[export]{adjustbox}
\usepackage[noend]{algpseudocode}
\usepackage{caption}
\usepackage{float}
\usepackage{booktabs}
\usepackage{tabularx}
\usepackage{makecell}
\usepackage{pdfpages}
\usepackage{enumitem}
\usepackage{amsmath,amssymb}
\usepackage{color}
\usepackage{subfig}
\usepackage{url}
\usepackage{subfig}
\usepackage{natbib}

\newcommand{\etal}{\mbox{\emph{et al.}}}

\begin{document}

\title{TextTopicNet - Self-Supervised Learning of Visual Features Through Embedding Images on Semantic Text Spaces}
\titlerunning{TextTopicNet}        

\author{Yash Patel           \and
        Lluis Gomez          \and
        Raul Gomez           \and
        Mar\c{c}al Rusi\~{n}ol \and
        Dimosthenis Karatzas \and
        C.V. Jawahar
}


\institute{Yash Patel \at
    The Robotics Institute, Carnegie Mellon University, PA, USA.\\
    CVIT, KCIS, IIIT Hyderabad, India.\\
    \email{yashp@andrew.cmu.edu}
    \and
    Lluis Gomez \at
    Computer Vision Center, Universitat Autonoma de Barcelona, Spain.\\
    \email{lgomez@cvc.uab.es}
    \and
    Raul Gomez  \at
    Eurecat, Centre Tecnològic de Catalunya, Barcelona, Spain.\\
    Computer Vision Center, Universitat Autonoma de Barcelona, Spain.\\
    \email{raul.gomez@cvc.uab.es}
    \and
    Mar\c{c}al Rusi\~{n}ol \at
    Computer Vision Center, Universitat Autonoma de Barcelona, Spain.\\
    \email{marcal@cvc.uab.es}
    \and
    Dimosthenis Karatzas \at
    Computer Vision Center, Universitat Autonoma de Barcelona, Spain.\\ 
    \email{dimos@cvc.uab.es}
    \and
    C.V. Jawahar \at
    CVIT, KCIS, IIIT Hyderabad, India.\\
    \email{jawahar@iiit.ac.in}
}

\date{Received: date / Accepted: date}

\maketitle

\begin{abstract}
 
The immense success of deep learning based methods in computer vision heavily relies on large scale training datasets. These richly annotated datasets help the network learn discriminative visual features. Collecting and annotating such datasets requires a tremendous amount of human effort and annotations are limited to popular set of classes. As an alternative, learning visual features by designing auxiliary tasks which make use of freely available self-supervision has become increasingly popular in the computer vision community. 

In this paper, we put forward an idea to take advantage of multi-modal context to provide self-supervision for the training of computer vision algorithms. We show that adequate visual features can be learned efficiently by training a CNN to predict the semantic textual context in which a particular image is more probable to appear as an illustration. More specifically we use popular text embedding techniques to provide the self-supervision for the training of deep CNN. 

Our experiments demonstrate state-of-the-art performance in image classification, object detection, and multi-modal retrieval compared to recent self-supervised or naturally-supervised approaches.

\keywords{Self-Supervised Learning \and Visual Representation Learning \and Topic-Modeling \and Multi-Modal Retrieval \and CNN}

\end{abstract}

\section{Introduction}
\label{sec:introduction}

The emergence of large-scale annotated datasets \citep{deng2009imagenet,zhou2014learning,lin2014microsoft} has undoubtedly been one of the key ingredients for the tremendous impact of deep learning on almost every computer vision task. However, there is a major issue with the supervised learning setup in large scale datasets: collecting and manually annotating those datasets requires great amount of human effort.

\begin{figure*}
\centering
\includegraphics[width=\textwidth, height=5cm]{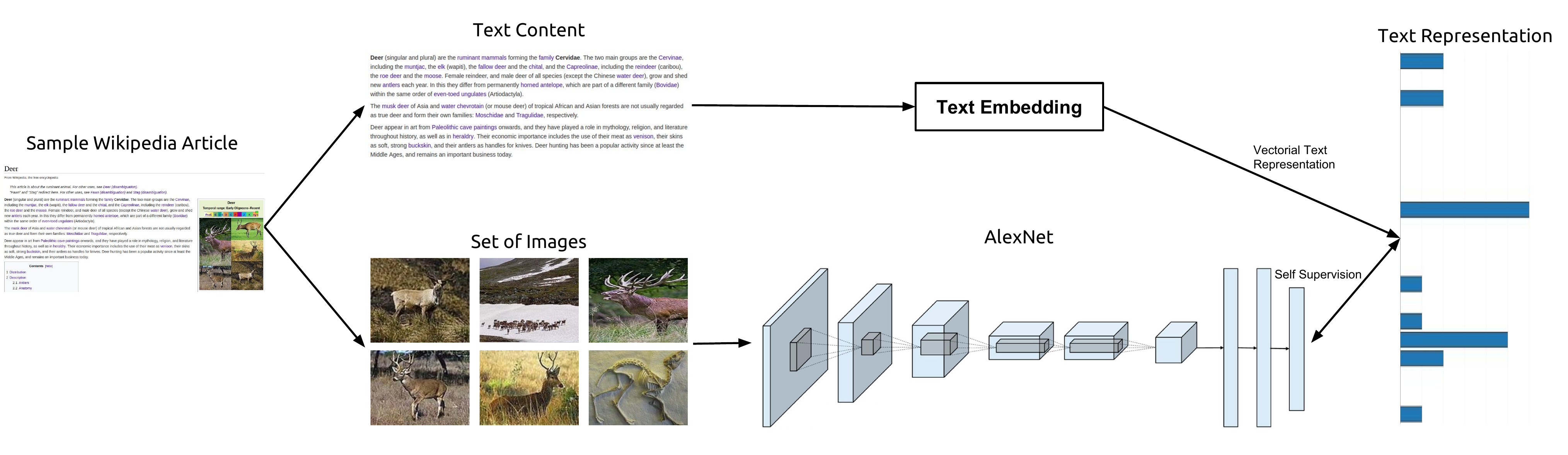}
\caption{Method overview: Wikipedia articles contain textual description of a subject, these articles are also accompanied with illustrative images supporting the text. A text embedding framework generates a global contextual representation of the textual information. This vectorial text representation of entire text article is used to provide the self-supervision for the training of CNN.}
\label{fig:overall_method}
\end{figure*}

As an alternative, self-supervised learning aims at learning discriminative visual features by designing auxiliary tasks where the target labels are free to obtain. These labels provide supervision for the training of computer vision models the same as in supervised learning, but could be directly obtained from the training data, either from the image itself~\citep{doersch2015unsupervised,pathak2016context} or from a complementary modality that is found naturally correlated with it~\citep{agrawal2015learning,owens2016ambient}. Unlike supervised learning that learns visual features from the human generated semantic labels, the self-supervised learning scheme mines them from the nature of the data. Another class of methods is weakly-supervised learning, where training makes use of low level human annotations for solving more complex computer vision tasks. One such example is making use of per-image class labels for object detection~\citep{bilen2016weakly,oquab2015object} in natural scene images.

In most cases, human generated data annotations consist of semantic entities in the form of textual information with different granularity depending on the vision task at hand: a single word to identify an object/place (classification), a list of words that describe the image (labeling), or a descriptive phrase of the scene shown (captioning). In this paper we propose that text found in illustrated articles can be leveraged as a form of image annotation to provide self-supervision, albeit being a very noisy one. The key benefit of this approach is that these annotations can be obtained for ``free''.

Illustrated articles are ubiquitous in our culture: for example in newspapers, encyclopedia entries, web pages, etc. Their visual and textual content complement each other and provide an enhanced semantic context to the reader. In this paper we propose to leverage all this freely available multi-modal content to train computer vision algorithms.

Surprisingly, the use of naturally co-occurring textual and visual information has not been fully utilized yet for self-supervised learning. The goal of this paper is to propose an alternative solution to fully supervised training of CNNs by leveraging the correlation between images and texts found in illustrated articles. Our main objective is to explore the strength of language semantics in unstructured text articles as a supervisory signal to learn visual features.

We present a method we call TextTopicNet, that performs self-supervised learning of visual features by mining a large scale corpus of multi-modal web documents (Wikipedia articles). TextTopicNet makes use of freely available unstructured multi-modal content for learning visual features in a self-supervised learning setup. 

We claim that it is feasible to learn discriminative features by training a CNN to predict the semantic context in which a particular image is more probable to appear as an illustration. As illustrated in Figure~\ref{fig:overall_method} our method consists in applying a text embedding algorithm to the textual part to obtain a vectorial text representation and then use this representation as the supervisory signal for visual learning of a CNN. We investigate the use of various document level and word level text embeddings of articles, and we empirically find that the the best practice is to represent the textual information at the topic level, by leveraging the hidden semantic structures discovered by the Latent Dirichlet Allocation (LDA) topic modeling framework~\citep{blei2003latent}.

\begin{figure}[h]
\centering
\subfigure[]{\includegraphics[width=0.23\textwidth]{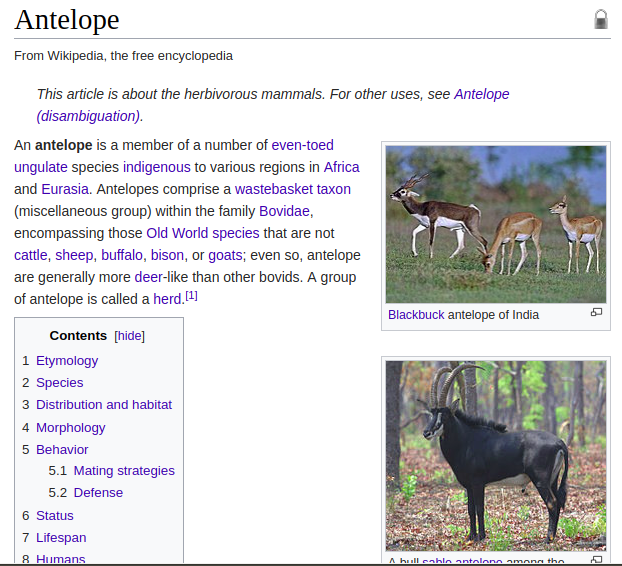}}
  \hfill
\subfigure[]{\includegraphics[width=0.23\textwidth]{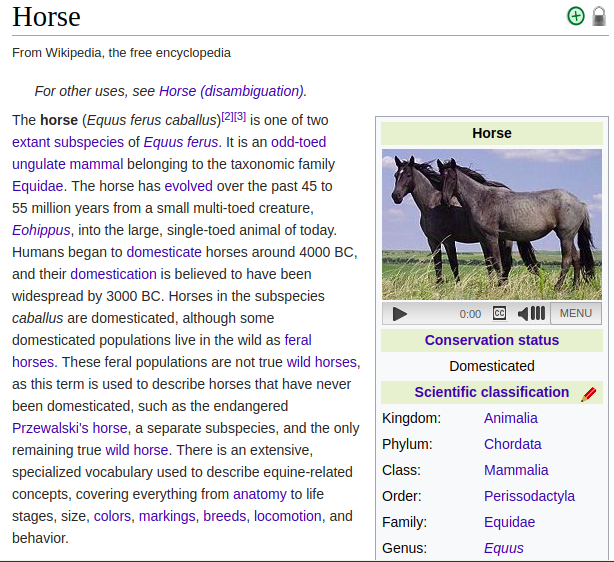}}\\
  \subfigure[]{\includegraphics[width=0.5\textwidth]{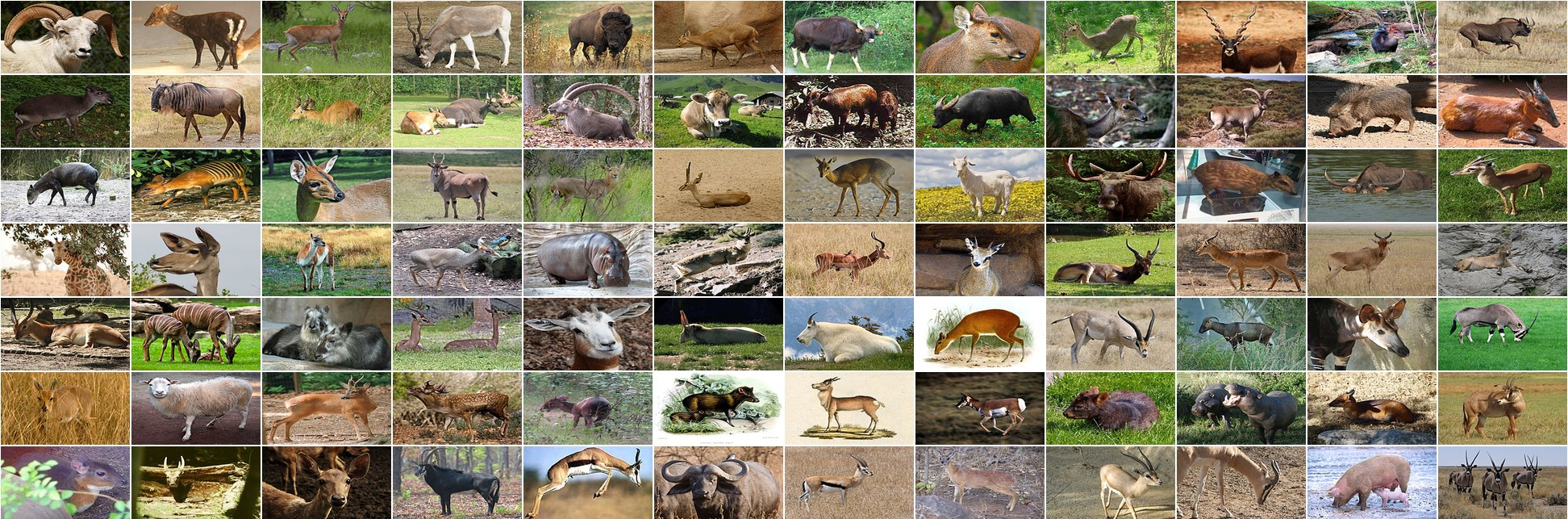}}
\caption{Illustrated Wikipedia articles about specific entities, like ``Antelope'' (a) or ``Horse'' (b), typically contain around five images. The total number of images for broader topics, e.g. ``herbivorous mammals'' (c), can easily reach hundreds or thousands.}
\label{fig:wiki_samples}
\end{figure}

As illustrated in Figure \ref{fig:wiki_samples}, the intuition behind using topic-level semantic descriptors is that the amount of visual data available about specific objects or fine-grained classes (e.g. a particular animal) is limited in our data collection, while it would be easy to find enough images representative of broader object categories (e.g. ``mammals''). As a result of this approach the expected visual features that we learn are generic for a given topic, but still useful for other, more specific, computer vision tasks.  

By training a CNN to directly project images into a textual semantic space, TextTopicNet is not only able to learn visual features from scratch without any annotated dataset, but it can also perform multi-modal retrieval in a natural way without requiring extra annotation or learning efforts.

This paper is an extended version of the work previously published in CVPR 2017~\citep{gomez2017self}. Following are the contributions in this paper:
\begin{itemize}
\item We provide an extension of our previous method \citep{gomez2017self} and show that the idea of self-supervised learning using illustrated articles is scalable and can be extended to a larger training dataset (such as the entire English Wikipedia).
\item We experimentally demonstrate that TextTopicNet outperforms recent self-supervised or naturally supervised methods on standard benchmark evaluations. We extend our previous analysis to the more challenging SUN397 \citep{xiao2010sun} dataset, where TextTopicNet substantially reduces the performance gap between self-supervised and supervised training on ImageNet~\citep{deng2009imagenet}.
\item We show that using textual context based representations for training helps the network to automatically learn semantic multi-modal retrieval. On the task of image-text retrieval, TextTopicNet outperforms unsupervised methods and shows competitive performance compared to supervised approaches without making use of any class specific information.
\item We provide a baseline comparison across different text-embeddings such as word2vec \citep{mikolov2013efficient}, GloVe \citep{pennington2014glove}, FastText \citep{joulin2016bag}, doc2vec \citep{le2014distributed} for the purpose of self-supervised learning.
\item We publicly release an image-text article co-occurring dataset which consists of 4.2 million images and is obtained from entire English Wikipedia.

\end{itemize}

The rest of the paper is structured as follows. In Section \ref{sec:rel_work}, previous work is reviewed. In Section \ref{sec:wiki_data} details of the training dataset and scrapping setup are given. TextTopicNet method is presented in Section \ref{sec:texttopicnet} and is evaluated in Section \ref{sec:experiments}. Finally, conclusions are drawn in Section \ref{sec:conclusion}.

\section{Related Work}
\label{sec:rel_work}

\subsection{Self-Supervised Learning}
Work in unsupervised data-dependent methods for learning visual features has been mainly focused on algorithms that learn filters one layer at a time. A number of unsupervised algorithms have been proposed to that effect, such as sparse-coding, restricted Boltzmann machines (RBMs), auto-encoders \citep{zhao2015stacked}, and K-means clustering~\citep{coates2010analysis,dundar2015convolutional,krahenbuhl2015data}. However, despite the success of such methods in several unsupervised learning benchmark datasets, a generic unsupervised method that works well with real-world images does not exist.

As an alternative to fully-unsupervised algorithms, there has recently been a growing interest in self-supervised or naturally-supervised approaches that make use of non-visual signals, intrinsically correlated to the image, as a form to supervise visual feature learning. 

Agrawal \etal~\citep{agrawal2015learning} draw inspiration from biological observation that the living organisms learned visual perception for the purpose of moving and interacting with the environment. They make use of egomotion information obtained by odometry sensors mounted on a vehicle. The agent, that is, the vehicle, can be considered as a moving camera. Thus, they train a network using contrastive loss
formulation \citep{mobahi2009deep} to predict the camera transformations between two image pairs.  

Wang \& Gupta \etal~\citep{wang2015unsupervised} make use of videos as training data and use relative motion of objects as supervisory signal for training.  Their general idea is that two image patches connected by a tracker may contain same object or object parts. The relative motion information is obtained by using a standard unsupervised tracking algorithm. A Siamese-triplet network is then trained using a ranking loss function.

In a further extension, Wang \& Gupta \etal~\citep{wang2017transitive} model two different variations: (a) inter-instance variations (two objects in the same class should have similar features) (b) intra-instance variations (viewpoint, pose, deformations, illumination, etc.). They generate a data graph over object instances with two kinds of edges: (a) different viewpoints of same object instance (b) same viewpoint of different object instances. Similar to \citep{wang2015unsupervised} they train a VGG-16 \citep{simonyan2014very} based Siamese-triplet network using a ranking loss function for the two different types of data-triplet-pairs.

Doersch \etal~\citep{doersch2015unsupervised} use spatial context such as relative position of patches within an image to make the network learn object and object parts. They make use of an unlabeled collection of images and train a network to predict the relative position of second patch given the first patch. Owens \etal~\citep{owens2016ambient} make use of sound as a modality to provide supervisory signal. They do so by training a deep CNN to predict a hand-crafted statistical summary of sound associated with a video frame.

Pathak \& Efros \etal~\citep{pathak2016context} take inspiration from auto-encoders and proposed a context-encoder. They train a network using a combination of L2 loss and adversarial loss to generate arbitrary image regions conditioned on their surrounding.

Bojanowski \& Joulin \etal~\citep{bojanowski2017unsupervised} present an approach for unsupervised learning of visual features using Noise As Target (NAT) label for training. Their approach is domain agnostic and makes use of fixed set of target labels for training. They make use of  stochastic batch reassignment strategy and a separable square loss function.

In this paper we explore a different modality, text, for self-supervision of CNN feature learning. As mentioned earlier, text is the default choice for image annotation in many computer vision tasks. This includes classical image classification~\citep{deng2009imagenet,everingham2010pascal},  annotation~\citep{duygulu2002object,huiskes2008mir}, and captioning~\citep{Ordonez2011im2text,lin2014microsoft}. In this paper, we extend this to a larger level of abstraction by capturing text semantics with topic models. Moreover, we avoid using any human supervision by leveraging the correlation between images and text in a largely abundant corpus of illustrated web articles.

\subsection{Deep Learning Image-Text Embeddings}

Joint image and text embeddings have been lately an active research area. The possibilities of learning together from different kinds of data have motivated this field of study, where both general and applied research has been done. 
DeViSE \citep{frome2013devise} proposes a pipeline that, instead of learning to predict ImageNet classes, it learns to infer the Word2Vec \citep{mikolov2013efficient} representations of their labels. The result is a model that makes semantically relevant predictions even when it makes errors, and generalizes to classes outside of its labeled training set.

A similar idea is explored in the work of Gordo \& Larlus \etal ~\citep{gordo2017}, where image captions are leveraged to learn a global visual representation for semantic retrieval. They use a \textit{tf-idf} based BoW representation over the image captions as a semantic similarity measure between images and they train a CNN to minimize a margin loss based on the distances of triplets of query-similar-dissimilar images. 

Wang \etal~\citep{wang2016learning} propose a method to learn a joint embedding of images and text for image-to-text and text-to-image retrieval, by training a neural network to embed in the same space Word2Vec \citep{mikolov2013efficient} text representations and CNN extracted features. 

Other than semantic retrieval, joint image-text embeddings have also been used in more specific applications. Gordo \etal~\citep{gordo2015lewis} embed word images in a semantic space relying in the graph taxonomy provided by WordNet to perform text recognition. In a more specific application, Salvador \etal~\citep{salvador2017learning} propose a joint embedding of food images and its recipes to identify ingredients, using Word2Vec \citep{mikolov2013efficient} and LSTM representations to encode ingredient names and cooking instructions and a CNN to extract visual features from the associated images.

Our work differ from the previous image-text embedding methods in that we aim to learn generic and discriminative features in a self-supervised fashion without making use of any annotated dataset. 

\subsection{Topic Modeling}

\begin{figure*}
  \includegraphics[width=\textwidth]{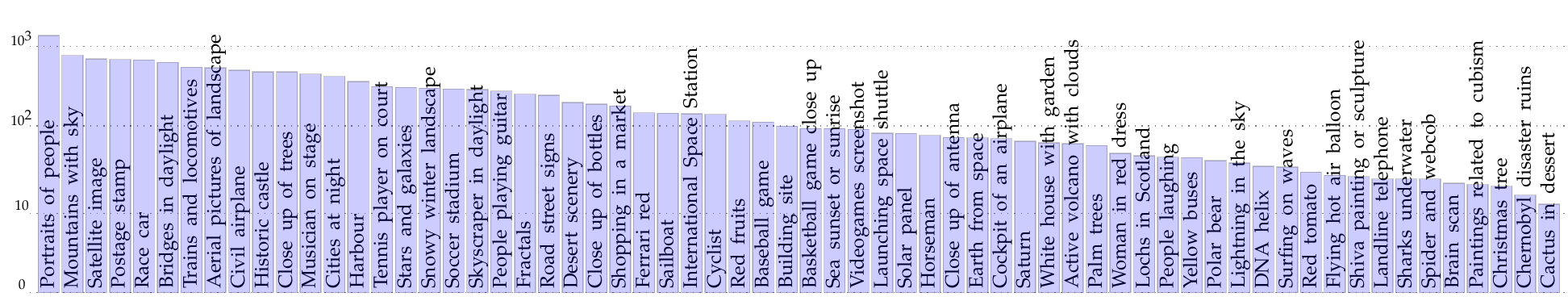}
    \caption{Number of relevant images (log scale) for a variety of semantic queries on the ImageCLEF Wikipedia collection~\citep{tsikrika2011overview}.}
  \label{fig:dataset_analysis}
\end{figure*}

Our method is also related with various image retrieval and annotation algorithms that make use of a topic modeling framework in order to embed text and images in a common space. Multi-modal LDA (mmLDA) and correspondence LDA (cLDA)~\citep{blei2003modeling} methods learn the joint distribution of image features and text captions by finding correlations between the two sets of hidden topics. Supervised variations of LDA are presented in ~\citep{rasiwasia2013latent,wang2011max,putthividhy2010topic} where the discovered topics are driven by the semantic regularities of interest for the classification task. Sivic \etal~\citep{sivic2005discovering} adopt BoW representation of images for discovering objects in images using pLSA~\citep{hofmann2001unsupervised} for topic modelling. Feng \etal~\citep{feng2010topic} uses the joint BoW representation of text and image for learning LDA.

Most cross-modal retrieval methods work with the idea of representing data of different modalities into a common space where data related to same topic of interest tend to appear together. The unsupervised methods in this domain utilize co-occurrence information to learn a common representation across different modalities. Verma \etal~\citep{verma2014im2text} do image-to-text and text-to-image retrieval using LDA~\citep{blei2003latent} for data representation. Methods such as those presented in ~\citep{rasiwasia2010new,gong2014multi,pereira2014role,li2011face} use Canonical Correlation Analysis (CCA) for establishing relationships between data of different modalities. Rasiwasia \etal~\citep{rasiwasia2010new} proposed a method for cross-modal retrieval by representing text using LDA~\citep{blei2003latent}, image using BoW and CCA for finding correlation across different modalities.

In one of our prior publication~\citep{patel2016dynamic}, we presented an approach for dynamic lexicon generation to improve scene text recognition systems. We used image-captions of MS-COCO \citep{lin2014microsoft} dataset and fine-tune an ImageNet \citep{deng2009imagenet} pre-trained Inception network \citep{szegedy2015going} to predict topic probabilities of a LDA model \citep{blei2003latent} directly from the images. Then using the word probabilities from LDA model, we predicted the probability of occurrence of each word given an image.

Our proposed method is related to these image annotation and image retrieval methods in the way that we use LDA~\citep{blei2003latent} topic-probabilities as common representation for both image and text. However, we differ from all these methods in that we use the topic level representations of text to supervise the visual feature learning of a Convolutional Neural Network. Our CNN model, by learning to predict the semantic context in which images appear as illustrations, learns generic visual features that can be leveraged for other visual tasks. 

\section{Wikipedia Image-Text Data}
\label{sec:wiki_data}

TextTopicNet leverages the semantic correlation of image-text pairs for self-supervised learning of visual features. Thus it requires a large scale dataset of multimodal content. In this paper we propose to use the Wikipedia web site as the source of such dataset. 

Wikipedia is a multilingual, web-based encyclopedia project currently composed of over 40 million articles across 299 different languages. Wikipedia articles are usually comprised by text and other kinds of multimedia objects (image, audio, and video files), and can thus be treated as multimodal documents.

For our experiments we make use of two different sets of Wikipedia articles' collections: (a) the ImageCLEF 2010 Wikipedia collection ~\citep{tsikrika2011overview} (b) our own contributed dataset Wikipedia Image-Text Co-ocurrence that is made publicly available and consists of $4.2$ million image-text pairs obtained from entire English Wikipedia.

\subsection{ImageCLEF Wikipedia Collection}

The  ImageCLEF  2010  Wikipedia  collection~\citep{tsikrika2011overview} consists of $237,434$ Wikipedia images and the Wikipedia articles that contain these images. 
An important observation is that the data collection and filtering is not semantically driven. The original ImageCLEF dataset contains all Wikipedia articles which have versions in three languages (English, German and French) and are illustrated with at least one image in each version. Thus, we have a broad distribution of semantic subjects, expected to be similar as to the entire Wikipedia or other general-knowledge data collections. A semantic analysis of the data, extracted from the ground-truth of relevance assessments for the ImageCLEF retrieval queries, is shown in Figure~\ref{fig:dataset_analysis}.
Although the dataset also provides human-generated annotations in this paper we train CNNs from scratch using only the raw Wikipedia articles and their images.

We consider only the English articles of the ImageCLEF Wikipedia collection. We also filter small images ($< 256$ pixels) and images with formats other than JPG (Wikipedia stores photographic images as JPG, and uses other formats for digital-born imagery). This way our subset of ImageCLEF training dataset is composed of $100,785$ images and $35,582$ unique articles. Throughout the paper, we refer to this dataset as ``ImageCLEF''.

\subsection{Full English Wikipedia dump}

In order to show that the idea of self-supervised learning using illustrated articles is scalable and can be extended to larger datasets than the ImageCLEF collection we have built a new dataset by scraping the entire English Wikipedia. With $5,614,418$ articles, the English Wikipedia is the largest among the 290 different Wikipedia encyclopedias.

\begin{figure}
\includegraphics[width=\linewidth]{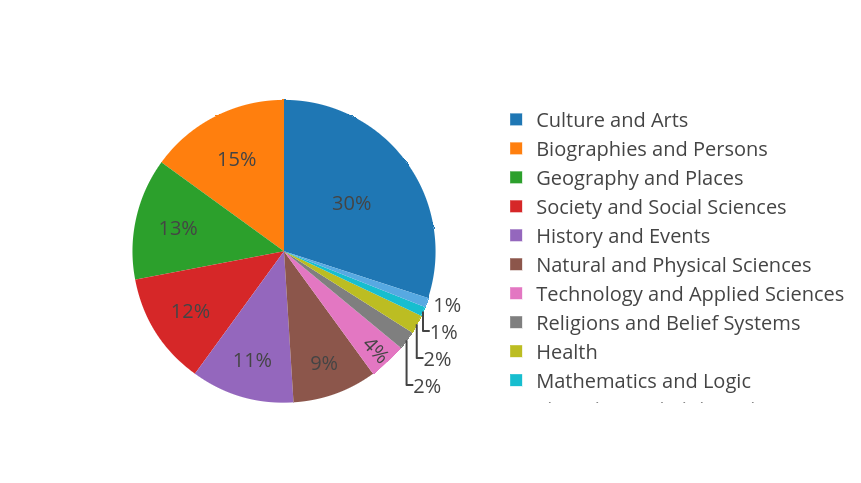}
\caption{Distribution of articles among the $11$ top level categories in the English Wikipedia~\citep{kittur2009s} computed from the page-category assignments.}
\label{fig:enwiki_topics}
\end{figure}

While a proper semantic analysis of the Wikipedia content is out of the scope of this paper, we consider relevant to highlight its broad and highly extensive coverage of human knowledge. For this, in Figure~\ref{fig:enwiki_topics} we show the distribution of articles among the $11$ top level categories as computed with the algorithm proposed by Kittur~\etal~\citep{kittur2009s}.

In order to obtain the training dataset for TextTopicNet we scrap entire English Wikipedia but we consider only articles with at least $50$ words and illustrated with at-least one image. Similarly to the preprocessing of ImageCLEF dataset we filter small images ($< 256$ pixels) and images with formats other than JPG. This way our training data is composed of $4.2$ million images and $1.7$ million unique articles, made publicly available\footnote{https://github.com/lluisgomez/TextTopicNet}. On average each text article is illustrated with $2.3$ images. Through rest of the paper, we refer to this dataset as ``Wikipedia''.

\begin{figure*}
  
  \includegraphics[width=\textwidth]{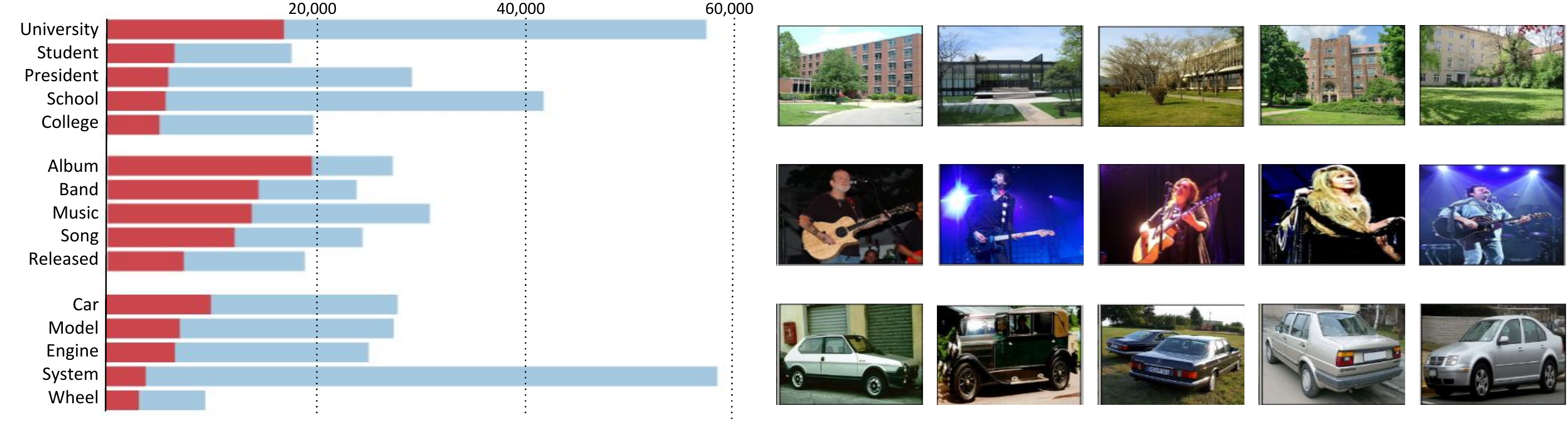}
    \caption{Top-5 most relevant words for 3 of the discovered topics by LDA analysis (left) on the ImageCLEF Wikipedia collection, and top-5 most relevant images for the same topics (right). Overall word frequency is shown in blue, and estimated word frequency within the topic in red.}
  \label{fig:topic_words_imgs}
\end{figure*}

\section{TextTopicNet}
\label{sec:texttopicnet}

The proposed method learns visual features in a self-supervised fashion by predicting the semantic textual context in which an image is more probable to appear as an illustration. As illustrated in Figure \ref{fig:overall_method} our CNN is trained on images to directly predict the vectorial representation of their corresponding text documents.

In Section \ref{sec:exp_compate_text_emb} we experimentally investigate the effect of various document level and word level text embeddings of articles for providing the training supervision to CNN. We provide a baseline comparison on the use of: Word2Vec \citep{mikolov2013efficient}, GloVe \citep{pennington2014glove}, FastText \citep{joulin2016bag}, Doc2Vec \citep{le2014distributed} and LDA \citep{blei2003latent} for the purpose of self-supervised learning. In this experiment we observe that using Latent Dirichlet Allocation (LDA) for text representation demonstrates best performance. 

On average, a Wikipedia article contains few hundred words and thus averaging word level representations such as Word2Vec \citep{mikolov2013efficient} looses semantic meaning. On the other hand LDA \citep{blei2003latent} discovers the distribution of documents over latent topics. Given a text document, this underlying distribution gives us a better semantic representation of entire text article.

In this section we discuss the specific details of the TextTopicNet pipeline using LDA for representing text articles. First, we describe how we learn a Latent Dirichlet Allocation (LDA) \citep{blei2003latent} topic model on all the text documents in our dataset. Then we detail how the LDA topic model is used to generate the target labels for training our Convolutional Neural Network (CNN).

\subsection{Latent Dirichlet Allocation (LDA) topic modeling}
\label{sec:method_lda}
Our self-supervised learning framework assumes that the textual information associated with the images in our dataset is generated by a mixture of hidden topics. Similar to various image annotation and image retrieval methods we make use of the Latent Dirichlet Allocation (LDA) \citep{blei2003latent} algorithm for discovering those latent topics and representing the textual information associated with a
given image as a probability distribution over the set of discovered topics. 

Representing text at topic level instead of at word level (BoW) provides us with: (1) a more compact representation (dimensionality reduction), and (2) a more semantically meaningful interpretation of descriptors.

LDA is a generative statistical model of a text corpus where each document can be viewed as a mixture of various topics, and each topic is characterized by a probability distribution over words. LDA can be represented as a three level hierarchical Bayesian model. Given a text corpus consisting of $M$ documents and a dictionary with $N$ words, LDA define the generative process for a document $d$ as follows:
\begin{itemize}
\item{Choose $\theta \sim Dirichlet(\alpha)$.}
\item{For each of the $N$ words $w_n$ in $d$:}
 \begin{itemize}
 \item{Choose a topic $z_n \sim Multinomial(\theta)$.}
 \item{Choose a word $w_n$ from $P(w_n \mid z_n, \beta)$, a multinomial probability conditioned on the topic $z_n$.}
 \end{itemize}
\end{itemize}
\noindent
where $\theta$ is the mixing proportion and is drawn from a Dirichlet prior with parameter $\alpha$, and both $\alpha$ and $\beta$ are corpus level parameters, sampled once in the process of generating a corpus. Each document is generated according to the topic proportions $z_{1:K}$ and word probabilities over $\beta$. The probability of a document $d$ in a corpus is defined as : 

\begin{equation}
P(d\mid\alpha, \beta) =  \nonumber
\int_{\theta}P(\theta \mid\alpha)\left(\prod_{n=1}^{N}\sum_{z_{K}}^{ } P(z_{K} \mid \theta)P(w_{n}\mid z_{K},\beta)\right)d\theta \nonumber
\end{equation}
\normalsize
Learning LDA on a document corpus provides two sets of parameters: word probabilities given topic $P(w\mid z_{1:K})$ and topic probabilities given document $P(z_{1:K} \mid d)$. Therefore each document is represented in terms of topic probabilities $z_{1:K}$ (being $K$ the number of topics) and word probabilities over topics. Any new (unseen) document can be represented in terms of a probability distribution over the topics of the learned LDA model by projecting it into the topic space.

\subsection{Self Supervised Learning of Visual Features using LDA Topic Probabilities}
\label{sec:method_cnn}
We train a CNN to predict text representations (topic probability distributions) from images. Our intuition is that we can learn useful visual features by training the CNN to predict the semantic context in which a particular image is more probable to appear as an illustration.

For our experiments we make use of two different architectures. One is the 8 layers CNN CaffeNet~\citep{jia2014caffe}, a replication of the AlexNet~\citep{krizhevsky2012imagenet} model with some differences (it does not train with the relighting data-augmentation, and the order of pooling and normalization layers is switched). The other architecture is a 6 layers CNN resulting from removing the 2 first convolutional layers from CaffeNet. This smaller network is used to do experiments with tiny images. The choice of AlexNet is justified because most of the existing self-supervised methods make use of this same architecture~\citep{owens2016ambient,agrawal2015learning,pathak2016context,wang2015unsupervised} and this makes us able to offer a direct comparison with them. 

For learning to predict the target topic probability distributions we minimize a sigmoid cross-entropy loss on our image dataset. We use a Stochastic Gradient Descent (SGD) optimizer, with base learning rate of  $0.001$, multiplied by $0.1$ every $250,000$ iterations, and momentum of $0.9$. The batch size is set to $128$. With these settings the network converges after $520,000$ iterations.

\begin{figure*}
\includegraphics[width=\textwidth]{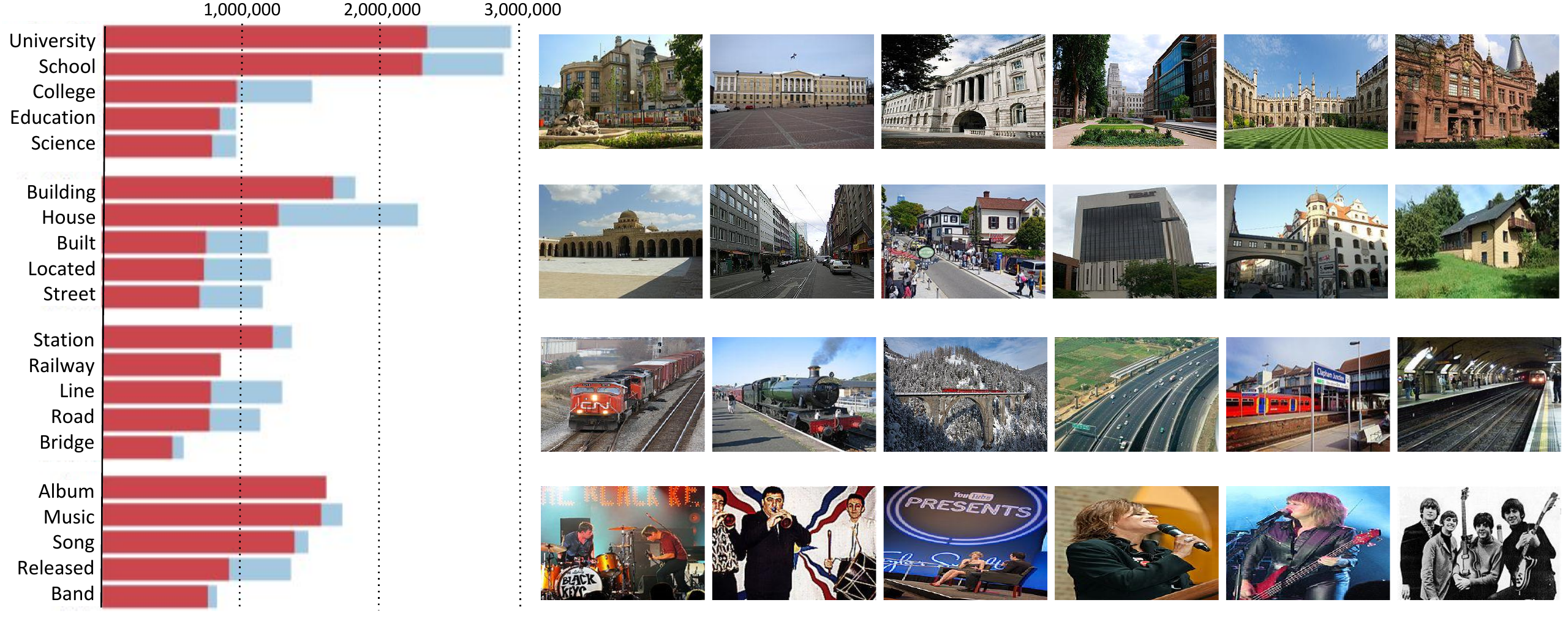}
\caption{Top-5 most relevant words for 4 of the discovered topics by LDA analysis (left) on the entire English Wikipedia dataset, and top-6 most relevant images for the same topics (right). Overall word frequency is shown in blue, and estimated word frequency within the topic in red.}
\label{fig:topic_vis}
\end{figure*}

In an attempt to visualize the semantic nature of the supervisory signal provided by the LDA model, we show in Figures~\ref{fig:topic_vis} and ~\ref{fig:topic_words_imgs} the top-5 most relevant words for discovered topics by LDA and the corresponding most relevant images for such topics. The analysis is done individually on each of the datasets introduced in Section~\ref{sec:wiki_data}.  We appreciate that the discovered topics correspond to broad semantic categories for which, a priori, it is difficult to find the most appropriate illustration. Still we observe that the most representative images for each topic present some regularities and thus allow the CNN to learn discriminative features for broader object classes, despite the possible noise introduced by outlier images that may appear in articles from the same topic. Further, by comparing the discovered topics from the ImageCLEF dataset (Figure \ref{fig:topic_words_imgs}) to the ones discovered in the Wikipedia dataset (Figure \ref{fig:topic_vis}) we can appreciate that the two LDA models share some common topics (eg. words like ``music'', ``album'', ``song'' are prominent to one of the topic in both LDA models). This observation supports the claim made in Section~\ref{sec:wiki_data} that both datasets must have a similar distribution of semantic subjects.

It is important to notice that a given image will rarely correspond to a single semantic topic, because by definition the discovered topics by LDA have a certain semantic overlap. In this sense we can think of the problem of predicting topic probabilities as a multi-label classification problem in which all classes exhibit a large intra-class variability. These intuitions motivate our choice of a sigmoid cross-entropy loss for predicting targets interpreted as topic probabilities instead of a one hot vector for a single topic.

Once the TextTopicNet model has been trained, it can be straightforwardly used in an image retrieval setting. Furthermore, it can be potentially extended to an image annotation \citep{patel2016dynamic} or captioning system by leveraging the common topic space in which text and images can be projected respectively by the LDA and CNN models.

However, in this paper we are more interested in analyzing the qualities of the visual features that we have learned by training the network to predict semantic topic distributions. We claim that the learned features, out of the common topic space, are not only of sufficient discriminative power but also carry more semantic information than features learned with other state of the art self-supervised and unsupervised approaches.

The proposed self-supervised learning framework will have thus a broad application in different computer vision tasks. With this spirit we propose the use of TextTopicNet as a convolutional feature extractor and as a CNN pre-training method. We evaluate these scenarios in the next section and compare the obtained results in different benchmarks with the state of the art.

\section{Experiments}
\label{sec:experiments}

In this section we perform extensive experimentation in order to demonstrate the quality of the visual features learned by the TextTopicNet model. 
Our aim is to demonstrate that the learned visual features are both discriminative and robust towards unseen or uncommon classes.

First we compare various text-embeddings for the purpose of self-supervised learning from pairs of images and texts. Second we perform a baseline analysis of TextTopicNet top layers' features for image classification on the PASCAL VOC 2007 dataset~\citep{everingham2010pascal} to find the optimal number of topics of the LDA model. Third we compare our method with state of the art self-supervised methods and unsupervised learning algorithms for image classification on PASCAL, SUN397~\citep{xiao2010sun}, and STL-10~\citep{coates2010analysis} datasets, and for object detection in PASCAL. Finally, we perform experiments on image retrieval from visual and textual queries on the Wikipedia retrieval dataset \citep{rasiwasia2010new}.

\subsection{Comparing Text-Embeddings for Self-Supervised Visual Feature Learning}
\label{sec:exp_compate_text_emb}

As we have previously mentioned in the review of the state of the art, there exist several text-image embedding pipelines that share the basic design of TextTopicNet but make use of other text representations instead of LDA topic probabilities. Thus our first objective is to understand the strength of using different text embeddings to provide self-supervision for CNN training. 

In order to do so, we train AlexNet \citep{krizhevsky2012imagenet} as explained in Section \ref{sec:method_cnn} on ImageCLEF dataset using different text embeddings: LDA \citep{blei2003latent}, Word2Vec \citep{mikolov2013efficient}, Doc2Vec \citep{le2014distributed}, GloVe \citep{pennington2014glove} and FastText \citep{joulin2016bag}. For evaluation of these trained models, we train one vs. rest SVMs using the image representation obtained by different layers on PASCAL VOC 2007 dataset \citep{everingham2010pascal}.

The PASCAL VOC 2007 dataset \citep{everingham2010pascal} consists of 9,963 images, split into 50\% for training/validation and 50\% for testing. Each image has been annotated with a bounding box and object class label for each object in one of the twenty classes present in the image: 
\emph{``person'', ``bird'', ``cat'', ``cow'', ``dog'', ``horse'', ``sheep'', ``aeroplane'', ``bicycle'', ``boat'', ``bus'', ``car'', ``motorbike'', ``train'', ``bottle'', ``chair'', ``dining table'', ``potted plant'', ``sofa'',} and \emph{``tv/monitor''}. The dataset is a standard benchmark for image classification and object detection tasks and its relatively small size for training  makes it specially well suited for the evaluation of self-supervised algorithms as well as for transfer learning methods.

Popular text vectorization methods in this pipelines are diverse in terms of architecture and the text structure they are designed to deal with. Some methods are oriented to generate representations of individual words \citep{joulin2016bag,mikolov2013efficient,pennington2014glove} and others to vectorize entire text articles or paragraphs \citep{blei2003latent,le2014distributed}. In our analysis we consider the top-performing text embeddings and test them in our pipeline to evaluate the performance of the learned visual features. 

Briefly the main characteristics of each text embedding method used in this experiment are as follows:
\begin{itemize}
\item \textbf{Word2Vec} \citep{mikolov2013efficient}: Using large amounts of unannotated plain text, Word2Vec learns relationships between words automatically using a feed-forward neural network. It builds distributed semantic representations of words using the context of them considering both words before and after the target word.

\item \textbf{Doc2Vec} \citep{le2014distributed}: Extends the Word2Vec idea to documents. Instead of learning feature representations for words, it learns them for sentences or documents.

\item \textbf{GloVe} \citep{pennington2014glove}: It is a count-based model. It learns the vectors by essentially doing dimensionality reduction on the co-occurrence counts matrix. Training is performed on aggregated global word-word co-occurrence statistics from a corpus.

\item \textbf{FastText} \citep{joulin2016bag}:  It is an extension of Word2Vec which treats each word as composed of character ngrams, learning representations for ngrams instead of words. The vector for a word is made of the sum of its character n grams, so it can generate embeddings for out of vocabulary words.
\end{itemize}

While LDA \citep{blei2003latent} and Doc2Vec \citep{le2014distributed} can directly generate text-article level embeddings, Word2Vec \citep{mikolov2013efficient}, GloVe \citep{pennington2014glove} and FastText \citep{joulin2016bag} generate only word level embeddings. In order to make use of representation obtained from entire text article for supervision, we use the mean embedding of all words within an article.

For all embeddings except LDA, we test with two different representation dimensions: (a) 40 (same as optimum number of topics when using LDA \citep{blei2003latent,gomez2017self}) (b) 300 (same as standard models as trained in original implementation). We make use of Gensim\footnote{\url{http://radimrehurek.com/gensim}} implementations of Word2Vec \citep{mikolov2013efficient}, FastText \citep{joulin2016bag} and Doc2Vec \citep{le2014distributed} and the GloVe implementation by provided by Maciej Kula\footnote{\url{http://github.com/maciejkula/glove-python}}.

For each of these text-embeddings we train a CNN as explained in Section \ref{sec:method_cnn}. Once the CNN is trained, we learn one-vs-all SVMs on features obtained from different layers in the network. Table~\ref{pascal_SVM_mAP_comparison} shows the PASCAL VOC2007 image classification performance using different text embeddings. We observe that Latent Drichilet Allocation (LDA) \citep{blei2003latent} based embedding serves as best global representation for self-supervised learning of visual features.

\begin{table}[h]
\centering
\begin{tabular}{ l | c | c | c }
\toprule
Text Representation &  pool5 & fc6 & fc7 \\
\midrule
LDA \citep{gomez2017self} & \textbf{47.4} & \textbf{48.1} & \textbf{48.5} \\
Word2Vec (40) & 44.1 & 45.1 & 36.9  \\
Word2Vec (300) & 41.1 & 36.6 & 32.2 \\
Doc2Vec (40) & 41.8 & 40.0 & 33.3 \\
Doc2Vec (300) & 43.7 & 35.4 & 33.1 \\
GloVe (40) & 41.6 & 40.6 & 34.7\\
GloVe (300) & 36.2 & 30.3 & 29.4\\
FastText (40) & 45.3 & 46.2 & 38.7\\
FastText (300) & 40.4 & 34.5 & 34.0\\
\bottomrule
\end{tabular}
\caption{TextTopicNet comparison using different text embeddings. PASCAL VOC2007 \%mAP image classification.}
\label{pascal_SVM_mAP_comparison}
\end{table}

\begin{table*}
\resizebox{\textwidth}{!}{
\begin{tabular}{l | c | c | c | c | c | c | c | c | c | c | c | c | c | c | c | c | c | c | c | c }
\toprule
Method &aer &bk &brd &bt &btl &bus &car &cat &chr &cow &din &dog &hrs &mbk &prs &pot &shp &sfa &trn &tv \\
\midrule
TextTopicNet (Wikipedia) &\textbf{71} &\textbf{52} &\textbf{47} &\textbf{61} &\textbf{26} &\textbf{49} &\textbf{71} & \textbf{46} &\textbf{47} &\textbf{36} &\textbf{44} &\textbf{41} &72 &\textbf{62} &\textbf{85} &\textbf{31} &\textbf{40} & \textbf{42} &\textbf{72} &\textbf{44}\\
TextTopicNet (ImageCLEF) &67 &44 &39 &53 &20 &49 &68 &42 &43 &33 &41 &35 &70 &57 &82 &30 &31 &39 &65 &41 \\
\midrule
Sound~\citep{owens2016ambient} &69 &45 &38 &56 &16 &47 &65 & 45 &41 &25 &37 &28 &\textbf{74} &61 &85 &26 &39 &32 &69 &38 \\
Texton-CNN &65 &35 &28 &46 &11 &31 &63 &30 &41 &17 &28 &23 &64 &51 &74 &9 &19 &33 &54 &30 \\
K-means &61 &31 &27 &49 &9 &27 &58 &34 &36 &12 &25 &21 &64 &38 &70 &18 &14 &25 &51 &25\\
Motion~\citep{wang2015unsupervised} &67 &35 &41 &54 &11 &35 &62 &35 &39 &21 &30 &26 &70 &53 &78 &22 &32 &37 &61 &34 \\
Patches~\citep{doersch2015unsupervised} &70 &44 &43 &60 &12 &44 &66 &52 &44 &24 &45 &31 &73 &48 &78 &14 &28 &39 &62 &43 \\
Egomotion~\citep{agrawal2015learning} &60 &24 &21 &35 &10 &19 &57 &24 &27 &11 &22 &18 &61 &40 &69 &13 &12 &24 &48 &28 \\
\midrule
ImageNet~\citep{krizhevsky2012imagenet} &79 &\textbf{71} &\textbf{73} &75 &\textbf{25} &60 &80 &\textbf{75} &51 &\textbf{45} &60 &\textbf{70} &\textbf{80} &\textbf{72} &\textbf{91} &42 &\textbf{62} &56 &82 &62 \\
Places~\citep{zhou2014learning} &\textbf{83} &60 &56 &\textbf{80} &23 &\textbf{66} &\textbf{84} &54 &\textbf{57} &40 &\textbf{74} &41 &\textbf{80} &68 &90 &\textbf{50} &45 &\textbf{61} &\textbf{88} &\textbf{63} \\
\bottomrule
\end{tabular}}
\caption{PASCAL VOC2007 per-class average precision (AP) scores for the classification task with pool5 features.}
\label{pascal_pool5_AP}
\end{table*}

\subsection{LDA Hyper-parameter Settings}
\label{sec:exp_lda_params}

As observed in Section \ref{sec:exp_compate_text_emb}, LDA \citep{blei2003latent} based global text representation of entire text articles provide best supervision. Here we perform a baseline analysis for parameter optimization using the standard train/validation split of the PASCAL VOC 2007 dataset. 

In this experiment we train a LDA topic model on the corpus of $35,582$ articles from the ImageCLEF Wikipedia collection~\citep{tsikrika2011overview}. From the raw articles we remove stop-words and punctuation, and perform lemmatization of words. The word dictionary ($50,913$ words) is made from the processed text corpus by filtering those words that appear in less than $20$ articles or in more than $50\%$ of the articles. At the time of choosing the number of topics in our model we must consider that as the number of topics increase, the documents of the training corpus are partitioned into finer collections, and increasing the number of topics may also cause an increment on the model perplexity~\citep{blei2003latent}. Thus, the number of topics is an important parameter in our model.

\begin{figure}[h!]
\includegraphics[width=0.5\textwidth]{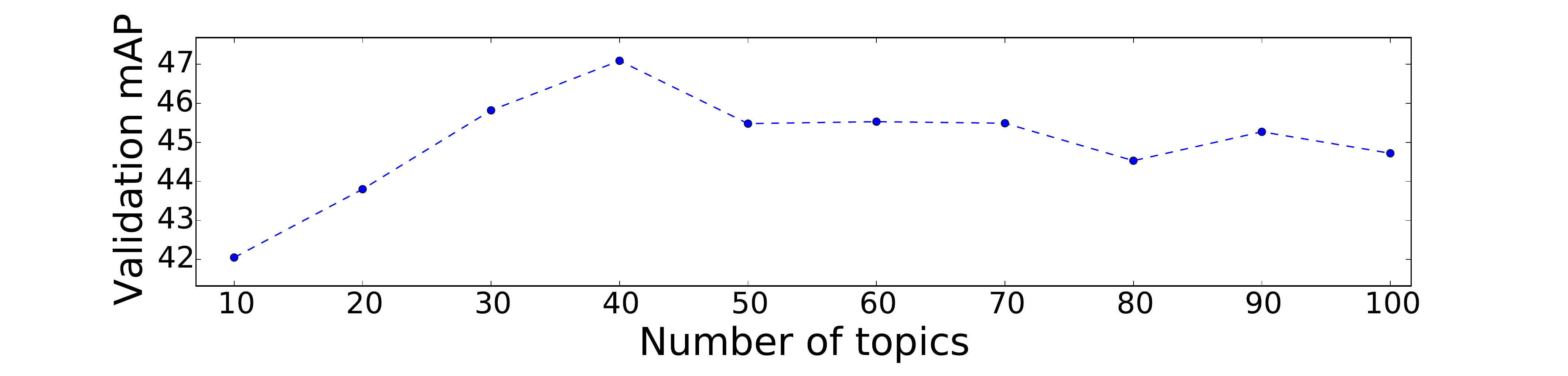}
\caption{One vs. Rest linear SVM validation \%mAP on PASCAL VOC2007 by varying number of topics of LDA~\citep{blei2003latent} in our method.}
  \label{plot_map_num_topics}
\end{figure}

We take a practical approach and empirically determine the optimal number of topics in our model by leveraging validation data. Figure~\ref{plot_map_num_topics} shows validation accuracy of SVM classification using \textit{fc7} features for different number of topics in our model. We appreciate that the best validation performance is obtained with the $40$ topics LDA model. This configuration is kept for both LDA models on ImageCLEF and Wikipedia datasets for the rest of our experiments. We do not perform this hyper-parameter optimization of LDA on introduced entire Wikipedia dataset due to high training time on $4.2$ million images.

\subsection{Image Classification}
\label{sec:exp_classification}

In this set of experiments we evaluate how good are the learned visual features of the 6 layer CNN (CaffeNet) for image classification when trained with the self-supervised method explained in Section~\ref{sec:texttopicnet}. Image classification is evaluated using two standard protocols: (1) training one-vs-all SVMs on representations obtained from different layers such as \textit{conv5, fc6, fc7} (2) fine-tuning the network on different datasets using TextTopicNet initialized weights.

\subsubsection{Unsupervised Features for Image Classification}  

Starting from the TextTopicNet model trained as detailed on Section~\ref{sec:texttopicnet} we extract features from the top layers of the CNN (fc7, fc6, pool5, etc.) for each image of the dataset. Then, for each class we perform a grid search over the parameter space of an one-vs-all Linear SVM classifier~\footnote{Liblinear implementation from \url{http://scikit-learn.org/}} to optimize its validation accuracy. Then, we use the best performing parameters to train again the one-vs-all SVM using both training and validation images.

Tables~\ref{pascal_pool5_AP} and~\ref{pascal_SVM_mAP} compare our results on the PASCAL test set with different state-of-the-art self-supervised learning algorithms using features from different top layers and SVM classifiers. Scores for all other methods are taken from~\citep{owens2016ambient}. We appreciate in Table~\ref{pascal_SVM_mAP} that using text semantics as supervision for visual feature learning outperforms all other modalities in this experiment. 

In Table~\ref{pascal_pool5_AP}, attention is drawn to the fact that our \textit{pool5} features are substantially more discriminative than the rest for the most difficult classes, see e.g. ``bottle'', ``pottedplant'' or ``cow''. Indeed, in the case of ``bottle'' our method outperforms fully supervised networks. Additionally for commonly occurring classes such as ``aeroplane'', ``car'', ``person'' TextTopicNet substantially outperforms previous self-supervised approaches and show competitive performance to supervised training.

\begin{table}[h]
\begin{tabular}{ l | c | c | c | c}
\toprule
Method & max5 & pool5 & fc6 & fc7 \\
\midrule
TextTopicNet (Wikipedia) & - &   \textbf{51.9} &  \textbf{54.2} & \textbf{55.8}\\
TextTopicNet (ImageCLEF) & - & 47.4 & 48.1 & 48.5 \\
\midrule
Sound ~\citep{owens2016ambient} & 39.4 & 46.7 & 47.1 & 47.4 \\
Texton-CNN & 28.9 & 37.5 & 35.3 & 32.5 \\
K-means~\citep{krahenbuhl2015data} & 27.5 & 34.8 & 33.9 & 32.1 \\
Tracking~\citep{wang2015unsupervised} & 33.5 & 42.2 & 42.4 & 40.2 \\
Patch pos.~\citep{doersch2015unsupervised} & 26.8 & 46.1 & - & - \\
Egomotion~\citep{agrawal2015learning} & 22.7 & 31.1 & - & - \\
\midrule
TextTopicNet (MS-COCO) & - & \textbf{50.7} & \textbf{53.1} & \textbf{55.4} \\
\midrule
ImageNet~\citep{krizhevsky2012imagenet} & \textbf{63.6} & \textbf{65.6} & \textbf{69.6} & \textbf{73.6} \\
Places~\citep{zhou2014learning} & 59.0 & 63.2 & 65.3 & 66.2 \\
\bottomrule
\end{tabular}
\caption{PASCAL VOC2007 \%mAP for image classification.}
\label{pascal_SVM_mAP}
\end{table}

TextTopicNet (Wikipedia) and TextTopicNet (ImageCLEF) in Table~\ref{pascal_SVM_mAP} correspond to the models trained respectively on each of the datasets detailed in Section~\ref{sec:wiki_data}. We appreciate that our model greatly benefits from the larger scale of the entire Wikipedia dataset. The TextTopicNet (COCO) entry corresponds to a model trained with MS-COCO~\citep{lin2014microsoft} images and their ground-truth caption annotations as textual content. Since MS-COCO captions are generated by human annotators, this entry can not be considered a self-supervised method, but rather as a kind of weakly supervised approach. Our interest in training this model is to show that having more specific textual content, like image captions, helps TextTopicNet to learn better features. In other words, there is an obvious correlation between the noise introduced in the self supervisory signal of our method and the quality of the learned features. Actually, the ImageNet entry in Table~\ref{pascal_SVM_mAP} can be seen as a model with a complete absence of noise, i.e. each image corresponds exactly to one topic and each topic corresponds exactly to one class (a single word). Still, the TextTopicNet (Wikipedia) features, learned from a very noisy signal, surprisingly outperform the ones of the TextTopicNet (COCO) model.

As an additional experiment we have calculated the 
classification performance on the combination of TextTopicNet and that of Sound entries in Table~\ref{pascal_SVM_mAP}. Here we seek insight about how complementary are the features learned with two different modalities of supervisory signals. By using the concatenation of \textit{fc7} features from TextTopicNet(ImageCLEF) and Sound models the mAP increases to 54.81\%. On combining \textit{fc7} features from TextTopicNet(Wikipedia) and Sound models the mAP gets to 57.38\%. This improvement in performance indicates towards a certain degree of complementarity.

Table~\ref{tab:sun_SVM_mAP} compares our results on the SUN397 \citep{xiao2010sun} test set with state-of-the-art self-supervised learning algorithms. SUN397 \citep{xiao2010sun} consists of 50 training and 50 test images for each of the 397 scene classes. We follow the same evaluation protocol as \citep{owens2016ambient,agrawal2015learning} and make use 20 images per class for training and remaining 30 for validation. We evaluate TextTopicNet on three different partitions of training and testing and report the average performance. This scene classification dataset is suitable for the evaluation of self-supervised approaches as it contains less frequently occurring classes and thus is more challenging compared to PASCAL VOC 2007 dataset.

We appreciate that TextTopicNet outperforms all other modalities of supervision in this experiment. We observe that using features from fc6 layer of TextTopicNet gives better performance compared to using features from fc7 layer. This indicates that fc6 and pool5 layers of TextTopicNet are more robust towards uncommon classes.

\begin{table}[h]
\begin{tabular}{ l | c | c | c | c}
\toprule
Method & max5 & pool5 & fc6 & fc7 \\
\midrule
TextTopicNet (Wikipedia) & - &  \textbf{28.8} &   \textbf{32.2} & \textbf{27.7} \\
\midrule
Sound ~\citep{owens2016ambient} & 17.1 &  22.5 & 21.3 & 21.4  \\
Texton-CNN & 10.7 & 15.2 & 11.4 & 7.6  \\
K-means~\citep{krahenbuhl2015data} & 11.6 & 14.9 & 12.8 & 12.4 \\
Tracking~\citep{wang2015unsupervised} & 14.1 & 18.7 & 16.2 & 15.1 \\
Patch pos.~\citep{doersch2015unsupervised} & 10.0 & 22.4 & - & - \\
Egomotion~\citep{agrawal2015learning} & 9.1 & 11.3 & - & -  \\
\midrule
ImageNet~\citep{krizhevsky2012imagenet} &  29.8 & 34.0 & 37.8 & 37.8 \\
Places~\citep{zhou2014learning} & \textbf{39.4} & \textbf{42.1} & \textbf{46.1} & \textbf{48.8}  \\
\bottomrule
\end{tabular}
\caption{SUN397 accuracy for image classification.}
\label{tab:sun_SVM_mAP}
\end{table}

\subsubsection{Self-Supervised pre-training for Image Classification}

In knowledge transfer, other that using CNN as a feature extractor and SVMs for classification, another standard procedure to evaluate the quality of CNN visual features it to fine-tune the network into the target domain. We analyze the performance of TextTopicNet for image classification by fine-tuning the CNN weights to specific datasets (PASCAL and STL-10). 

For fine-tuning our network we use the following optimization strategy:  we use Stochastic Gradient
Descent (SGD) for $120,000$ iterations with an initial learning rate of $0.0001$ (reduced by $0.1$ every $30,000$ iterations), batch size of $64$, and momentum of $0.9$. We use data augmentation by random crops and mirroring. At test time we follow the standard procedure of averaging the net responses at $10$ random crops. 

Table~\ref{pascal_finetuning_mAP} compares our results for image classification on PASCAL by fine-tuning the weights learned with different self-supervised learning algorithms. Image classification using AlexNet when trained only on PASCAL VOC dataset with randomly initialized weights achieve a performance of $53.4$ \%mAP. We appreciate that TextTopicNet substantially improved the classification performance over this baseline.

\begin{table}[h]
\centering
\begin{tabular}{l | c}
\toprule
Method & Fine-tuning  \\ 
\midrule
TextTopicNet (Wikipedia) & \textbf{61.0} \\
TextTopicNet (ImageCLEF)~\citep{gomez2017self} & 55.7 \\
\midrule
K-means~\citep{krahenbuhl2015data} & 56.6 \\
Tracking~\citep{wang2015unsupervised} & 55.6 \\
Patch pos.~\citep{doersch2015unsupervised} & 55.1 \\
Egomotion~\citep{agrawal2015learning} & 31.0 \\
NAT~\citep{bojanowski2017unsupervised} & 56.7 \\
Context Encoder~\citep{pathak2016context} &  56.5\\
\midrule
ImageNet\citep{krizhevsky2012imagenet} & \textbf{78.2}\\
\bottomrule
\end{tabular}
\caption{Fine-tuning results on PASCAL VOC 2007.}
\label{pascal_finetuning_mAP}
\end{table}

Table~\ref{stl10_finetuning_Acc} compares our classification accuracy on STL-10 with different state of the art unsupervised learning algorithms. In this experiment we make use of the shortened 6 layers network in order to adapt better to image sizes for this dataset ($96\times96$ pixels). We do fine-tuning with the same hyper-parameters as for the 6 layer network.

The standard procedure on STL-10 is to perform unsupervised training on a provided set of $100,000$ unlabeled images, and then supervised training on the labeled data. While our method does not directly compare with unsupervised and semi-supervised methods in Table~\ref{stl10_finetuning_Acc}, because of the distinct approach (self-supervision), the experiment  provides insight about the added value of self-supervision compared with fully-unsupervised data-driven algorithms. It is important to notice that we do not make use of the STL-10 unlabeled data in our training.

\begin{table}[h]
\centering
\begin{tabular}{ l | c  }
\hline
Method &  Acc. \\
\midrule
TextTopicNet (ImageCLEF) - CNN-finetuning *& \textbf{76.51\%} \\
TextTopicNet (ImageCLEF) - fc7+SVM *& 66.00\% \\
\midrule
Semi-supervised auto-encoder~\citep{zhao2015stacked} & \textbf{74.33}\% \\
Convolutional k-means~\citep{dundar2015convolutional} & 74.10\% \\
CNN with Target Coding~\citep{yang2015deep} & 73.15\% \\
Exemplar convnets~\citep{dosovitskiy2014discriminative} & 72.80\%\\
Unsupervised pre-training~\citep{paine2014analysis} & 70.20\% \\
Swersky \etal~\citep{swersky2013multi} * & 70.10\%\\
C-SVDDNet~\citep{wang2016unsupervised} & 68.23\% \\ 
K-means (Single layer net)~\citep{coates2010analysis} & 51.50\% \\
Raw pixels & 31.80\% \\
\bottomrule
\end{tabular}
\caption{STL-10 classification accuracy. Methods with an asterisk mark make use of external (unlabeled) data.}
\label{stl10_finetuning_Acc}
\end{table}

\subsubsection{Visual Features Analysis}

We further analyze the qualities of the learned features by visualizing the receptive field segmentation of TextTopicNet convolutional units using the methodology of~\citep{zhou2014object,owens2016ambient}. The purpose of this experiment is to gain insight in what our CNN has learned to detect.

\begin{figure}[h]
  \includegraphics[width=\columnwidth]{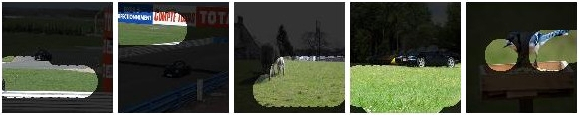}\\
  \includegraphics[width=\columnwidth]{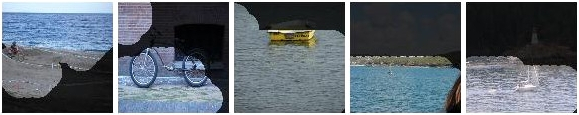}\\
  \includegraphics[width=\columnwidth]{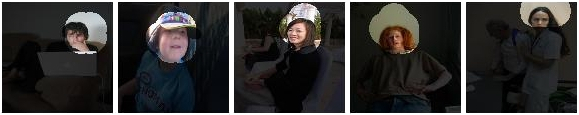}\\
  \includegraphics[width=\columnwidth]{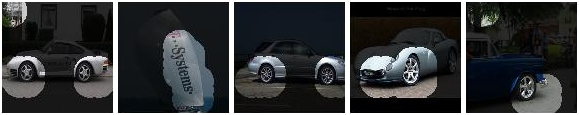}\\
  \includegraphics[width=\columnwidth]{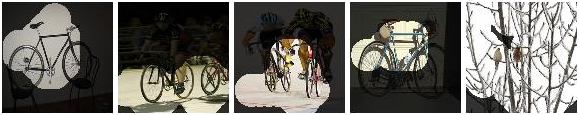}
  \caption{Top-5 activations for five units in \textit{fc7} layer of TextTopicNet(ImageCLEF) model. While most TextTopicNet units are selective to generic textures, like grass or water, some of them are also selective for specific shapes, objects, and object-parts.}
  \label{rf_segmentations}
\end{figure}

Figure \ref{rf_segmentations} shows a selection of neurons in the \textit{fc7} layer of our model. We appreciate that our network units are quite generic, mainly selective to textures, shapes and object-parts, although some object-selective units are also present (e.g. faces).

\subsection{Object Detection}
\label{sec:exp_detection}

Similar to other self-supervised approaches, for object detection we make use of Fast R-CNN~\citep{girshick2015fast}. We replace the ImageNet initialized weights of Fast R-CNN with the weights of TextTopicNet and train the network with default parameters for $40,000$ iterations on training and validation set of PASCAL VOC 2007.

Table~\ref{tab:pascal_finetuning_detection} compares our results for image classification and object detection on the test set of PASCAL VOC 2007 with different self-supervised learning algorithms.
Object detection using Fast RCNN \citep{girshick2015fast} when trained only on PASCAL VOC dataset with randomly initialized weights achieve a performance of $40.7$ \%mAP. We appreciate that TextTopicNet and other self-supervised methods enhance the detection performance over this baseline.

\begin{table}[h]
\centering
\begin{tabular}{ l | c }
\toprule
Method  & Detection \\
\midrule
TextTopicNet (Wikipedia) & 44.3 \\
TextTopicNet (ImageCLEF)  &  43.0 \\
\midrule
Sound ~\citep{owens2016ambient} & 44.1 \\
K-means~\citep{krahenbuhl2015data}  & 45.6 \\
Tracking~\citep{wang2015unsupervised}  & 47.4 \\
Patch pos.~\citep{doersch2015unsupervised}  & 46.6 \\
Egomotion~\citep{agrawal2015learning} & 41.8 \\
NAT~\citep{bojanowski2017unsupervised} & \textbf{49.4} \\
Context Encoder~\citep{pathak2016context} & 44.5\\
\midrule
ImageNet~\citep{krizhevsky2012imagenet} & \textbf{56.8} \\
\bottomrule
\end{tabular}
\caption{PASCAL VOC2007 finetuning \%mAP for object detection.}
\label{tab:pascal_finetuning_detection}
\end{table}

\subsection{Multi-Modal Retrieval}
\label{sec:exp_retrieval}

\begin{figure*}[t]
\includegraphics[width=\textwidth]{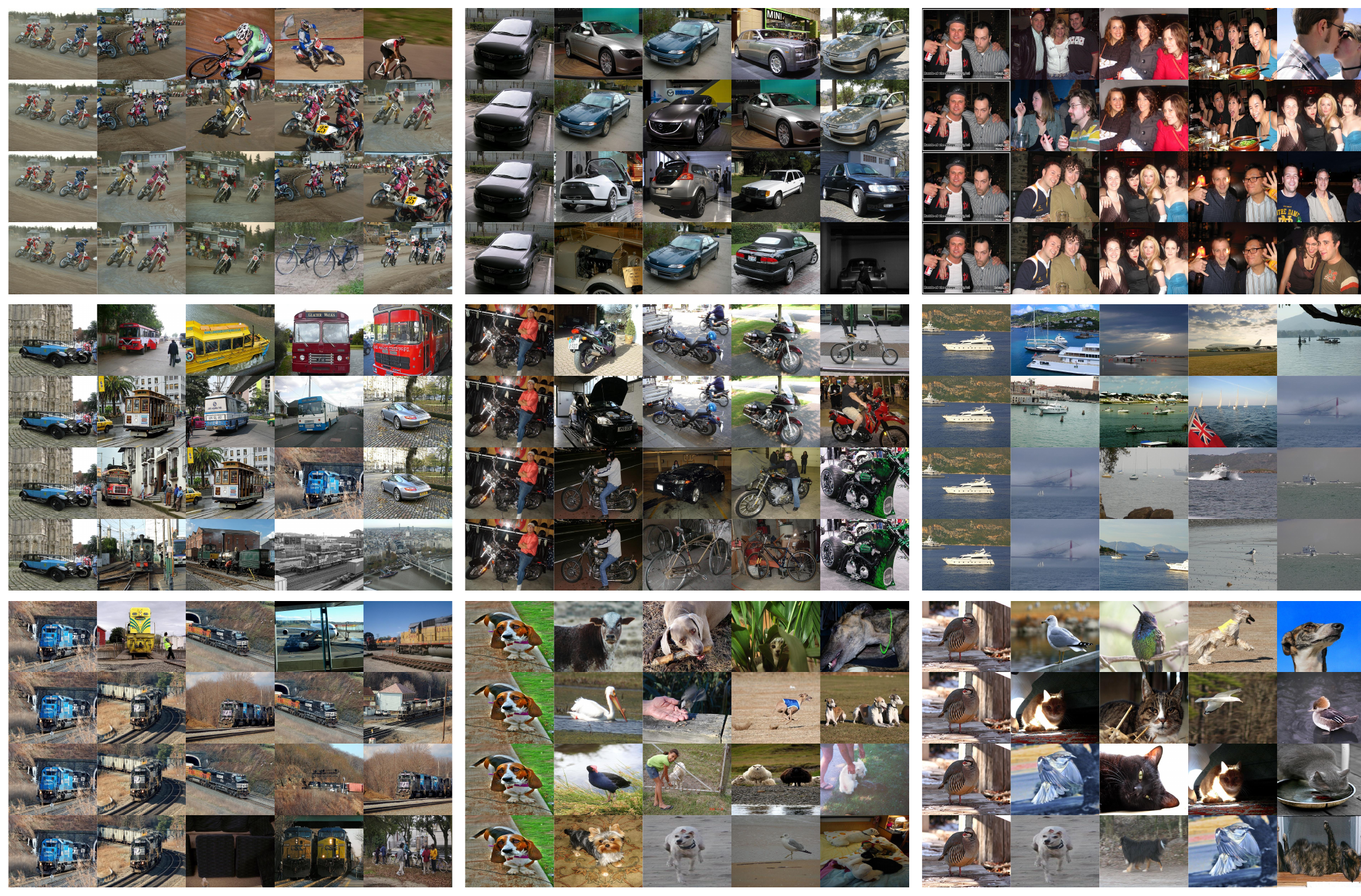}
\caption{Top 4 nearest neighbors for a given query image image (left-most). Each row makes use of features obtained from different layers of TextTopicNet (without fine tuning). From top to bottom: prob, fc7, fc6, pool5.}
\label{fig:img2img}
\end{figure*}

\begin{figure*}[t]
\includegraphics[width=\textwidth]{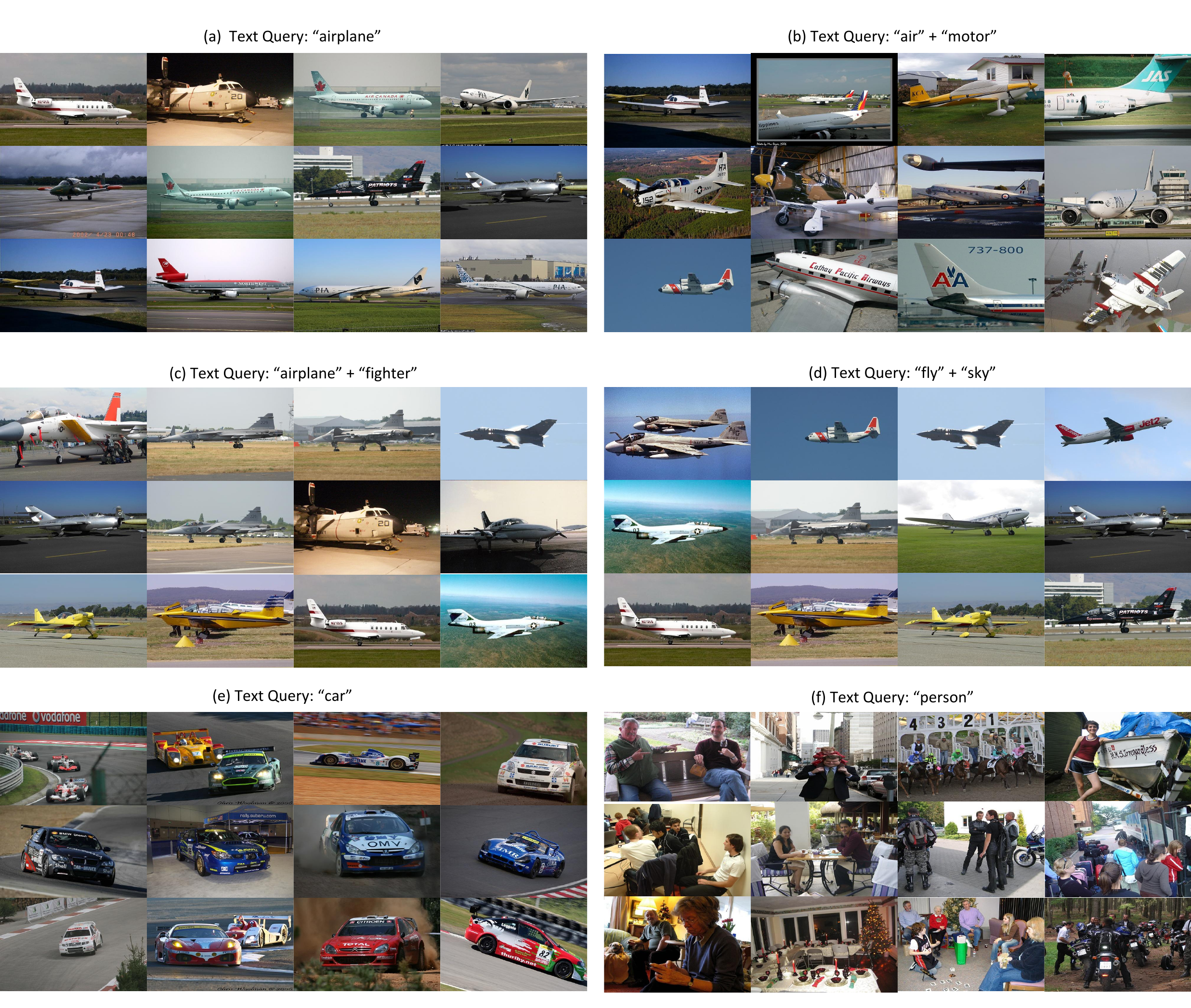}
\caption{Top 12 nearest neighbors for different text queries.}
\label{fig:txt2img}
\end{figure*}

\begin{figure}[H]
\includegraphics[width=0.50\textwidth]{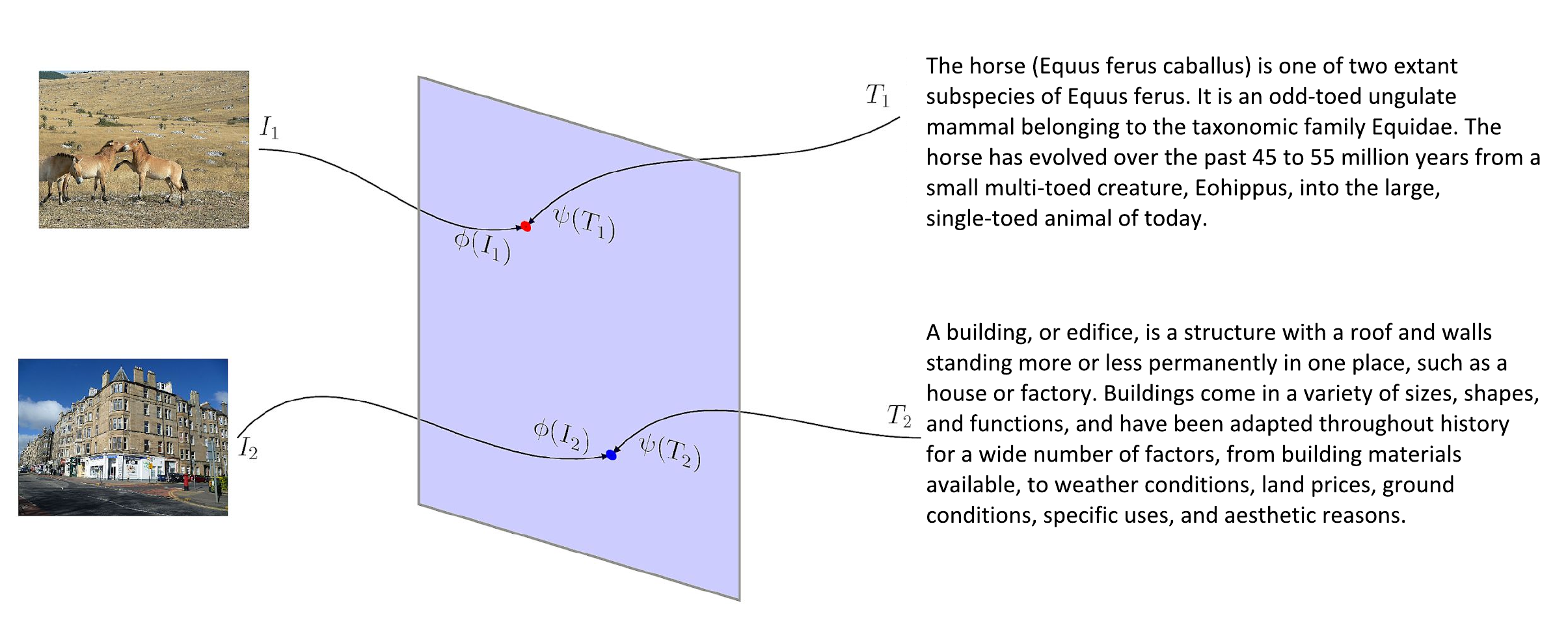}
\caption{TextTopicNet projects the images on same topic probability representation as that of co-occurring text article.}
\label{fig:multi_modal}
\end{figure}

We evaluate our learned self-supervised visual features for two types of multi-modal retrieval tasks: (1) Image query vs. Text database, (2) Text query vs. Image database. For this purpose, we use the Wikipedia retrieval dataset \citep{rasiwasia2010new}, which consists of 2,866 image-document pairs split into train and test set of 2,173 and 693 pairs respectively. Further, each image-document pair is labeled with one of ten semantic classes \citep{rasiwasia2010new}. As demonstrated in Figure \ref{fig:multi_modal} for retrieval we project images and documents into the learned topic space and compute the KL-divergence distance of the query (image or text) with all the entities in the database. 

In Table \ref{table:multi_modal_retrieval} we compare our results with supervised and unsupervised multi-modal retrieval methods discussed in \citep{wang2016comprehensive} and ~\citep{kang2015cross}. Supervised methods make use of class or categorical information associated with each image-document pair, whereas unsupervised methods do not. All of these methods use LDA for text representation and CNN features from pre-trained CaffeNet, which is trained on ImageNet dataset in a supervised setting. We appreciate that our self-supervised method outperforms unsupervised approaches, and has competitive performance to supervised methods without using any labeled data.

\begin{table}[h]
\centering
\begin{tabular}{l | c | c | c}
\toprule
Method & \shortstack{Image\\Query} & \shortstack{Text\\Query} & Average \\
\midrule
TextTopicNet (Wikipedia) & $37.63$ & $\textbf{40.25}$ & $38.94$\\
TextTopicNet (ImageCLEF) & $39.58$ & $38.16$ & $38.87$ \\
\midrule
CCA \citep{rasiwasia2010new}& $19.70$ & $17.84$ & $18.77$ \\
PLS \citep{rosipal2006overview} & $30.55$ & $28.03$ & $29.29$ \\
\midrule
SCM* \citep{rasiwasia2010new}& $37.13$ & $28.23$ & $32.68$ \\
GMMFA* \citep{sharma2012generalized} & $38.74$ & $31.09$ & $34.91$ \\
CCA-3V* \citep{gong2014multi} & $40.49$ & $36.51$ & $38.50$ \\
GMLDA* \citep{sharma2012generalized} & $40.84$ & $36.93$ & $38.88$ \\
LCFS* \citep{wang2013learning}& $41.32$ & $38.45$ & $39.88$ \\
JFSSL* \citep{wang2016joint}& $\textbf{42.79}$ & $39.57$ & $\textbf{41.18}$ \\
\bottomrule
\end{tabular}
\caption{MAP comparison on Wikipedia dataset \citep{rasiwasia2010new} with supervised (bottom) and unsupervised (middle) methods. Methods marked with asterisk make use of document (image-text) class category information.}
\label{table:multi_modal_retrieval}
\end{table}

Finally, in order to analyze better what is the nature of learned features by our self-supervised TextTopicNet we perform additional qualitative experiments for the image retrieval task.

Figure~\ref{fig:img2img} shows the 4 nearest neighbors for a given query image (left-most), where each row makes use of features obtained from different layers of TextTopicNet (without fine tuning). From top to bottom: prob, fc7, fc6, pool5. Query images are randomly selected from PASCAL VOC 2007 dataset and never shown at training time. It can be appreciated that when retrieval is performed in the topic space layer (prob, 40 dimensions, top row), the results are semantically close, although not necessarily visually similar. As features from earlier layers are used, the results tend to be more visually similar to the query image. Further we appreciate that without any supervision from PASCAL VOC 2007 classes, TextTopicNet learns to retrieve images belonging to correct corresponding class of input image.

Figure~\ref{fig:txt2img} shows the 12 nearest neighbors for a given text query in the topic space of TextTopicNet (again, without fine tuning). Interestingly, the list of retrieved images for the first query (``airplane'') is almost the same for related words and synonyms such as  ``flight'', ``airway'', or ``aircraft''. By leveraging textual semantic information our method learns a polysemic representation of images. Further it can be appreciated that TextTopicNet is capable of handling semantic text queries for retrieval such as (``airplane'' + ``fighter'', ``fly'' + ``sky'').

\section{Conclusions}
\label{sec:conclusion}

In this paper we provide an extension to our CVPR 2017 paper \citep{gomez2017self} on self-supervised learning using text topic spaces learned by LDA \citep{blei2003latent} topic model. The presented method, TextTopicNet, is able to take advantage of freely available multi-modal content to train computer vision algorithms without human supervision. By considering text found in illustrated articles as noisy image annotations the proposed method learns visual features by training a CNN to predict the semantic context in which a particular image is more probable to appear as an illustration. Here we experimentally demonstrate that our method is scalable to larger and more diverse training datasets. 

The contributed experiments show that although the learned visual features are generic for broad topics, they can be used for more specific computer vision tasks such as image classification, object detection, and multi-modal retrieval. Our results are superior when compared with state of the art self-supervised algorithms for visual feature learning.

{TextTopicNet source code, pre-trained models and introduced Wikipedia dataset (Section \ref{sec:wiki_data}) are publicly available at} \url{https://github.com/lluisgomez/TextTopicNet}.

\begin{acknowledgements}
This work has been partially supported by the Spanish research project TIN2014-52072-P, the CERCA Programme / Generalitat de Catalunya, the H2020 Marie Skłodowska-Curie actions of the European Union, grant agreement No 712949 (TECNIOspring PLUS), the Agency for Business Competitiveness of the Government of Catalonia (ACCIO), CEFIPRA Project 5302-1 and the project ``aBSINTHE - AYUDAS FUNDACI{\'O}N BBVA A EQUIPOS DE INVESTIGACION CIENTIFICA 2017. We gratefully acknowledge the support of the NVIDIA Corporation with the donation of the Titan X Pascal GPU used for this research.
\end{acknowledgements}

\bibliographystyle{spbasic}      
\bibliography{bibliography}

\end{document}